\pdfoutput=1
\documentclass[runningheads]{llncs}
\usepackage{graphicx}
\usepackage{amsmath,amssymb} % define this before the line numbering.
\usepackage{color}
\usepackage{array}
\usepackage{float}
\usepackage{subfig}
\usepackage{epsfig}
\usepackage{multirow}
\usepackage{bm}
\usepackage{cleveref}

\graphicspath{ {./figures/}
				{./figures/configuration/}
					}

\newcolumntype{C}[1]{>{\centering\let\newline\\\arraybackslash\hspace{0pt}}m{#1}}

\begin{document}
% \renewcommand\thelinenumber{\color[rgb]{0.2,0.5,0.8}\normalfont\sffamily\scriptsize\arabic{linenumber}\color[rgb]{0,0,0}}
% \renewcommand\makeLineNumber {\hss\thelinenumber\ \hspace{6mm} \rlap{\hskip\textwidth\ \hspace{6.5mm}\thelinenumber}}
% \linenumbers
%\pagestyle{headings}
%\mainmatter
%\def\ECCV18SubNumber{8}  % Insert your submission number here

\title{Brain-inspired robust delineation operator} % Replace with your title

\titlerunning{Brain-inspired robust delineation operator}

\authorrunning{Strisciuglio et al.}

\author{Nicola Strisciuglio \and George Azzopardi \and Nicolai Petkov}
\institute{Bernoulli Institute, University of Groningen, The Netherlands
\email{n.strisciuglio@rug.nl}}

\maketitle

\begin{abstract}
%Biologically-inspired approaches for computer vision and image processing aim at the implementation of algorithms that simulate certain properties of the visual system.  In particular, the capability of discerning objects also in very textured or noisy environments or the use of contextual and spatial relations between objects, are the source of our inspiration to design a new robust algorithm..

In this paper we present a novel filter, based on the existing COSFIRE filter, for the delineation of patterns of interest. It includes a mechanism of push-pull inhibition that improves robustness to noise in terms of spurious texture.  Push-pull inhibition is a phenomenon that is observed in neurons in area V1 of the  visual cortex, which suppresses the response of certain simple cells for stimuli of preferred orientation but of non-preferred contrast. This type of inhibition allows for sharper detection of the patterns of interest and improves the quality of delineation especially in images with spurious texture.

We performed experiments on images from different applications, namely  the detection of rose stems for automatic gardening, the delineation of cracks in pavements and road surfaces, and the segmentation of blood vessels in retinal images. Push-pull inhibition helped to improve results considerably in all applications.

\end{abstract}

\section{Introduction}
The delineation of elongated structures  is a fundamental process in image processing and computer vision, for various applications: detection of rose stems for automatic gardening robotics, crack delineation in roads or walls, segmentation of blood vessels in medical images, segmentation of roads and rivers in aerial images and so on.
In these applications, images usually contain large amounts of background noise and spurious texture, which cause segmentation errors~\cite{Zhang09river,Lacoste2005,StrisciuglioIWOBI17}. 

In this paper, we present a novel filter, inspired by the push-pull inhibition that is exhibited by some neurons in area V1 of the primary visual cortex. We construct a filter that has two components, an excitatory and an inhibitory one, based on the existing model of neurons with excitatory receptive fields in area V1, called CORF~\cite{AzzopardiCORF2012}, whose implementation is known as B-COSFIRE and shown to be effective for the delineation of blood vessels in medical images~\cite{Azzopardi2015,StrisciuglioVIP15}, also in combination with machin learnig techniques~\cite{Strisciuglio2016}.
We name the proposed filter RUSTICO, which stands for RobUST Inhibition-augmented Curvilinear Operator, and show how push-pull inhibition contributes to strenghten the robustness with respect to noise and spurious texture in the delineation of elongated patterns. The aim of this work is to demonstrate how inspiration from neurophysiological findings can be used to design effective algorithms, on which one can build more complex systems.

State-of-the-art approaches for the delineation of curvilinear patterns in images were recently reviewed in~\cite{Bibiloni2016}. Fundamental methods are based on a parametric formulation of the pattern of interest, namely line-like structures. The Hough transform, for instance, projects an input image onto a parameter space (slope and bias) in which linear segments are easier to detect. The disadvantage of parametric methods is that they require a strict mathematical model of the patterns of interest, and different shapes require different formulations.

Other methods are based on filtering or mathematical morphology, such as the Frangi detector that employs multi-scale local derivatives~\cite{Frangi1998}. Matched filters, which model the profile of the elongated patterns with 2D Gaussian kernels, were proposed in~\cite{Al-Rawi2007}. Combination of different techniques, such as Frangi filters and Gabor Wavelets, was also studied~\cite{Oliveira2016}. Mathematical morphology techniques assume \textit{a-priori} knowldege about the geometry of the patterns of interest~\cite{Mendonca2006}, such as size, orientation and width~\cite{MartinezPerez}, or concavity~\cite{Lam2010}. Recently, a method called RORPO, based on morphological path operators was proposed for the delineation of 2D and 3D curvilinear patterns~\cite{rorpoPAMI}. 

Point processes were also employed to segment networks of elongated structures, which are considered as complex sets of linear segments~\cite{Lacoste2005}. They were also combined with Gibbs models~\cite{Lafarge2010}, Monte-Carlo simulation~\cite{Verdi2012}, and graph-based representations with topological information about the line networks~\cite{Turetken16}. These methods are computationally very expensive, not being suitable for real-time or high resolution image processing.

Machine learning techniques have also been investigated to perform pixel-wise segmentation of elongated patterns. Early approaches in~\cite{Niemeijer2004} and~\cite{Soares2006} constructed pixel-wise feature vectors with multi-scale Gaussian and Gabor wavelet features, respectively, or with the responses of a bank of ridge detectors~\cite{StaalDrive2004}. Invariant moments were also used to describe the pixel characteristics in~\cite{Marin2011}.
More recently, Convolutional Neural Networks (CNNs) gained particular popularity in computer vision, for many tasks including segmentation. In~\cite{Liskowski2016}, for instance, image patches containing lines were used to train a CNN for the segmentation of blood vessels in medical images. More general architectures for segmentation were proposed, such as U-Net~\cite{Ronneberger2015} and SegNet~\cite{badrinarayanan2015segnet}. CNNs are supervised approaches, employ a large amount of filters at different stages and usually achieve high segmentation performance, but require large amounts of labeled training data to learn effective models and are computationally very expensive, requiring GPU hardware.

The approach that we introduce in this work is unsupervised and, hence, it is not appropriate to compare it with the performance of CNNs, but rather to demonstrate how we can make use of neuro-physiological evidence about the functioning of the visual system to improve image processing and computer vision algorithms. We demonstrate the effectiveness of the push-pull inhibition phenomenon and the improved robustness of the proposed filter in three applications were spurious textures are present, namely delineation of rose stems for garden robotics, detection of cracks in road surfaces and segmentation of blood vessels in retinal images. 

The paper is organized as follows. In Section~\ref{sec:method}, we describe the proposed implementation of the push-pull inhibition mechanism, and the data sets used for the experiments in Sections~\ref{sec:materials}. We present the results that we achieved, and provide a discussion of the results and comparison with those obtained by existing methods in Section~\ref{sec:experiments}. We draw conclusions in Section~\ref{sec:conclusions}.

\section{Method}
\label{sec:method}
The main idea of RUSTICO is the design of an operator selective for curvilinear patterns with push-pull inhibition. This type of inhibition is known to suppress responses to spurious texture and thus emphasizing more the detection of linear structures. In practice, RUSTICO takes input from two types of COSFIRE filters of the type introduced in \cite{Azzopardi2015}, one that gives excitatory input and the other that acts as inhibitory component. We compute the response of RUSTICO by subtracting the (weighted) response of the inhibitory component from the excitatory one.

\subsection{B-COSFIRE filter}
The COSFIRE filter approach is trainable, in that the selectivity of the filter is determined from an automatic configuration procedure that analyzes a given prototype pattern. In~\cite{Azzopardi2015}, a bar-like synthetic prototype pattern was used and the resulting filter responded strongly to blood vessels in retinal images. In~\cite{AzzopardiCORF2012}, instead, an edge prototype pattern was used for configuration of a COSFIRE filter that was very effective for contour detection.
%As shown in \cite{Azzopardi2015} when the pattern of interest, that is the prototype, is a bar-like structure, then the resulting B-COSFIRE (B for bar) filter responds in locations that are characterized by such patterns, and thus making the filter suitable for the delineation of vessel-like structures. Elsewhere, we demonstrated that if the prototype pattern is an edge \cite{AzzopardiCORF2012}, we can construct a filter that is very effective for contour detection, and when the prototype pattern has a more complicated shape, we can configure filters that are suitable for object localization and recognition \cite{fewpapers}.

\begin{figure}[!t]
\caption{(a) A prototype line and a (b) sketch of the configuration process.}
\centering
\setlength{\unitlength}{25mm}
\begin{tabular}{cc}
	\includegraphics[height=\unitlength]{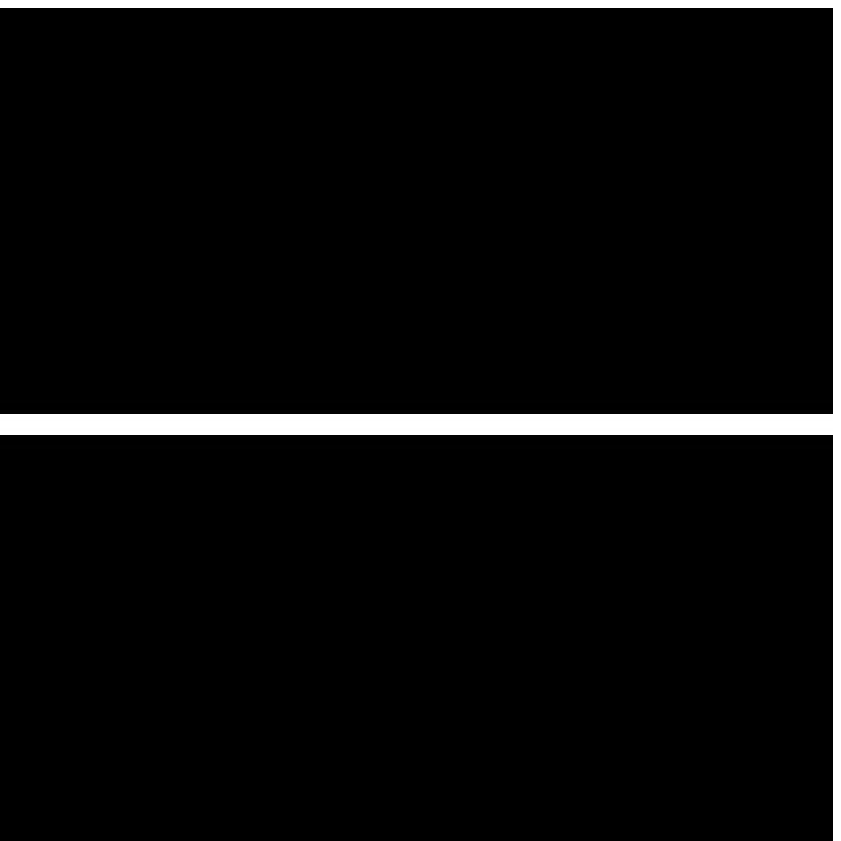} &
	\includegraphics[height=\unitlength]{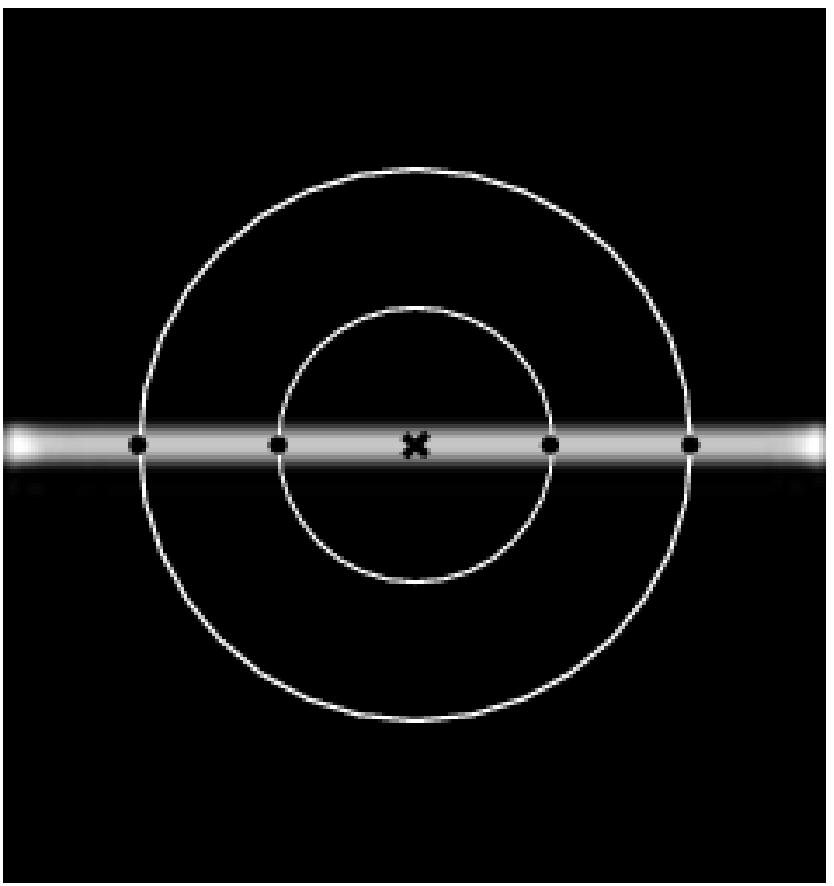} \\
	(a) & (b)
\end{tabular}
\label{fig:prototype}
\end{figure}

The automatic configuration of a COSFIRE filter involves two steps: first, the determination of keypoints in the given prototype pattern that we illustrate in Fig.~\ref{fig:prototype} by means of a system of concentric circles and linear filtering with difference-of-Gaussians (DoG) functions and secondly, the description of such keypoints in terms of four parameters. These parameters include the polarity $\delta$ and standard deviation $\sigma$, which describe the type (center-on or center-off) along with the spread of the outer Gaussian function\footnote{The standard deviation of the inner Gaussian function is $0.5\sigma$)} of the DoG that gives the maximum response in the concerned keypoint. The other two parameters are the distance $\rho$ and polar angle $\phi$ of the keypoint with respect to the center of the prototype. We define a COSFIRE filter as a set of 4-tuples $B= \{(\delta_i, \sigma_i, \rho_i, \phi_i)~|~ i=1\dots n\}$. %, where $\psi$ represents the orientation of the prototype, which is always assumed to be 0.

The response of a B-COSFIRE filter, denoted by $r_B(x,y)$, is computed by first calculating an intermediate feature map for each tuple followed by combining all features maps by geometric mean. The feature map of tuple $i$ is computed in four steps: firstly, \textit{convolve} the image with a DoG function whose polarity is $\delta_i$ and standard deviation $\sigma_i$, secondly \textit{rectify} the output by a rectification linear unit (ReLU) that sets to zero all negative responses, thirdly \textit{blur} the thresholded response map by a Gaussian function whose standard deviation $\sigma' = \sigma_0 + \alpha\rho_i$ where $\sigma_0$ and $\alpha$ are hyperparameters which we set empirically, and lastly \textit{shift} the blurred responses by the vector $[\rho_i,\pi-\phi_i]$. 

To configure a B-COSFIRE filter which is selective for the same prototype but rotated by a given offset $\psi$, one can construct a new filter $B_\psi$ by manipulating the angular parameter in all tuples of the filter $B$: $B^\psi = \{(\delta_i,\sigma_i,\rho_i,\phi_i+\psi) ~|~ \forall ~(\delta_i,\sigma_i,\rho_i,\phi_i) \in B\}$. This mechanism is required to achieve tolerance to rotation. For more details on COSFIRE filters we refer the reader to \cite{Azzopardi2015}. 

\subsection{Push-pull inhibition}
Push-pull inhibition is a phenomenon that has been observed in many simple cells~\cite{Taylor595}. It is thought that an interneuron inhibits the response of the simple cell to which it is connected. The receptive field of the interneuron is typically larger than that of the simple cell. The effect is that the response of the simple cell is suppressed for a stimulus with preferred orientation but with contrast opposite of the preferred one. We model the interneuron by another COSFIRE filter which we denote by $\hat{B}_\lambda$ and define it as $\hat{B}_\lambda = \{-\delta_i,\lambda\sigma_i,\rho_i,\phi_i~|~\forall ~(\delta_i,\sigma_i,\rho_i,\phi_i) \in B\}$ where $\lambda$ is a weighting factor  that controls the size of the afferent DoG functions. The response of the inhibitory COSFIRE filter $r_{\hat{B}_\lambda}(x,y)$ is computed with the same procedure described above.

\subsection{RUSTICO}
We define an orientation-selective RUSTICO as a pair $R_\lambda(B,\hat{B}_\lambda)$ and compute its response $R(x,y)$ by combining the excitatory and inhibitory inputs with a linear function:

\begin{equation}
	R(x,y) = |r_B(x,y) - \xi r_{\hat{B}_\lambda}(x,y)|^+
\end{equation}

\noindent where $\xi$ is the weighting or strength of the inhibitory component, which we determine experimentally, and $|.|^+$ indicates the ReLU function. 

In order to configure a RUSTICO filter that is tolerant to rotations we consider various pairs of excitatory and inhibitory COSFIRE filters that are selective for different orientations and then combine their response maps. Formally, we denote by $R^M_\Psi$ a multi-orientation RUSTICO and define it as a set $\tilde{R}_\Psi=\{(B^\psi,\hat{B}_\lambda^\psi)~|~\forall ~\psi \in \Psi\}$. Finally, the multi-orientation RUSTICO response $r_{\tilde{R}_\Psi}(x,y)$ is achieved by taking the maximum superposition of the response maps corresponding to all pairs in the set $\tilde{R}_\Psi$:

\begin{equation}
r_{\tilde{R}_\Psi}(x,y) = \max_{\psi \in \Psi}\{|r_{B^\psi}(x,y) - \xi r_{\hat{B}^{\psi}_\lambda}(x,y)|^+\}
\end{equation}

\section{Materials}
\label{sec:materials}
\begin{figure}[!t]
\caption{Example images from the considered data sets, together with the corresponding ground truth images.}
\centering
\setlength{\unitlength}{17mm}
\begin{tabular}{cccccc}
	\includegraphics[height=\unitlength]{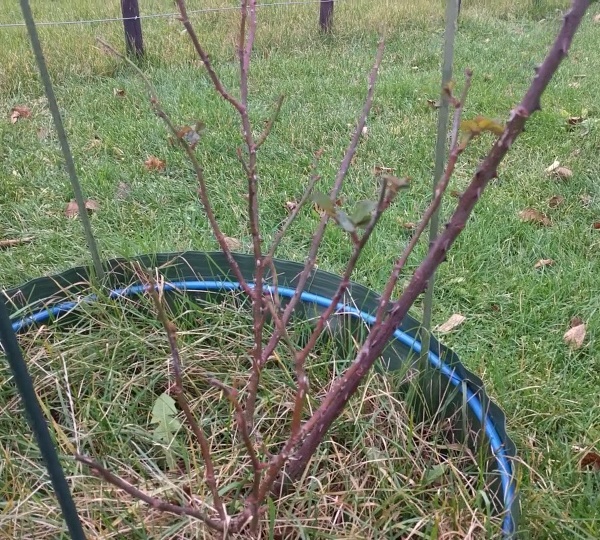} &
	\includegraphics[height=\unitlength]{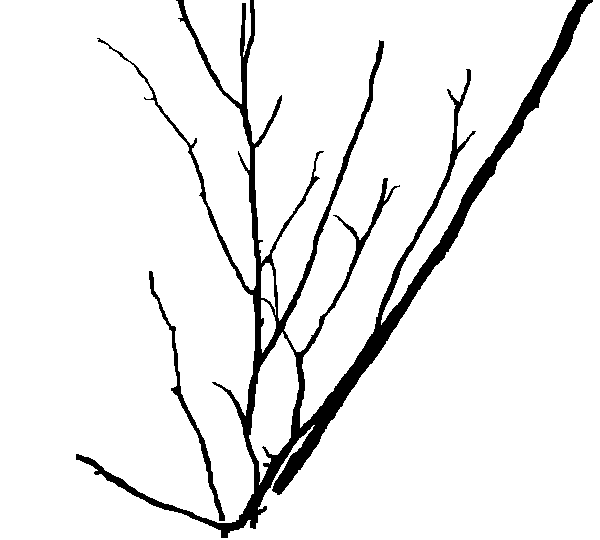} &
	\includegraphics[height=\unitlength]{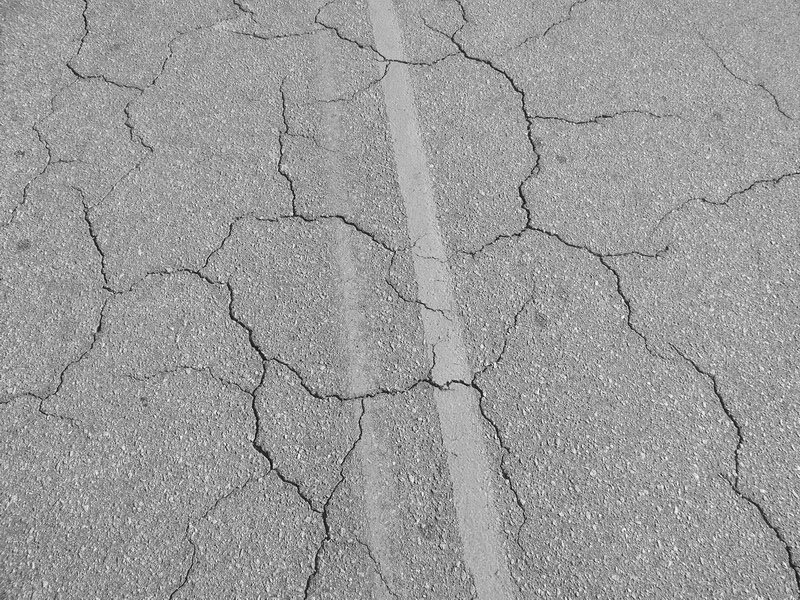} &
	\includegraphics[height=\unitlength]{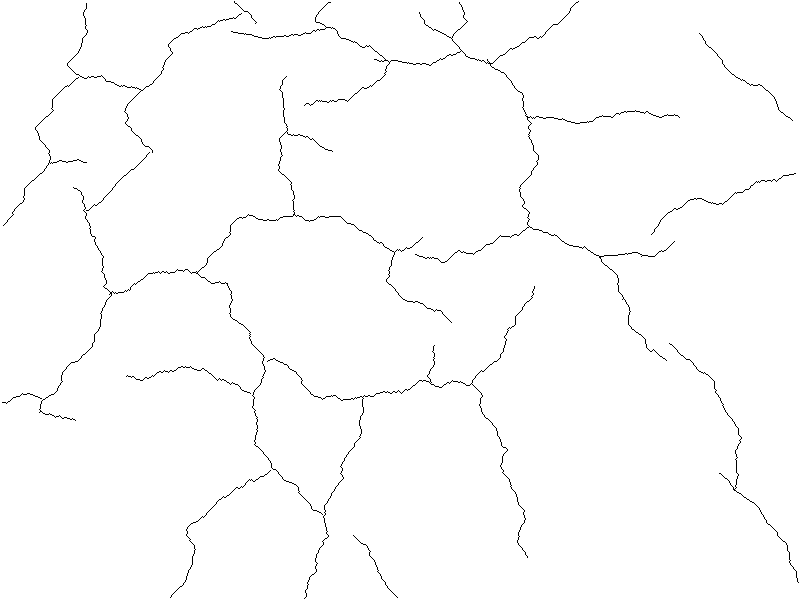} &
	\includegraphics[height=\unitlength]{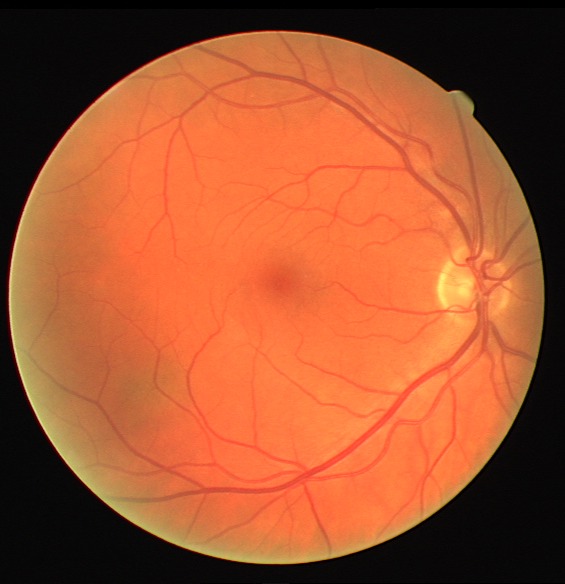} &
	\includegraphics[height=\unitlength]{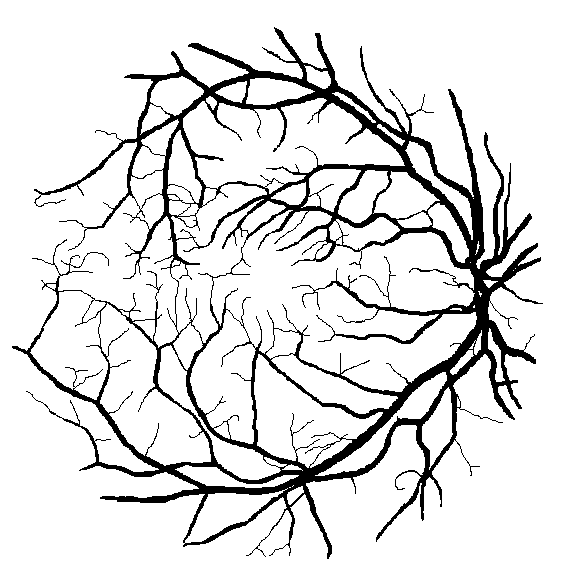}	
\end{tabular}
\label{fig:datasets}
\end{figure}

We tested the performance of the proposed operator with push-pull inhibition on different data sets, namely the TB-roses-1, the CrackTree206~\cite{Zou2012} and  the DRIVE~\cite{StaalDrive2004} data sets. We show example images from the three data sets, together with the corresponding ground truth delineation maps in Fig.~\ref{fig:datasets}.

The TB-roses-1 data set is composed of $100$ images, which we recorded in a real garden in the context of the TrimBot2020 project~\cite{trimbot}. It is designed for testing algorithms for delineation of rose branches in applications of gardening robotics. The images have resolution of $960 \times 540$ pixels and are provided together with two ground truth images, one indicating the centerline of the rose branches and the other marking the whole segmented branches. The data set is publicly available\footnote{http://gitlab.com/nicstrisc/RUSTICO}.

The CrackTree206 data set is composed of $206$ images of road surface, taken with an RGB camera at resolution $800\times 600$ pixels. The images contain spurious texture around the road cracks, due to the intrisic composition of the asphalt. This makes the delineation of cracks a hard task, since the size and contrast of the cracks are very similar to the one of the textured background. The images are provided with manually labeled images that delineate the center-line of the cracks and serve as ground truth for performance evaluation.

The DRIVE data set of retinal fundus images is divided into a training and a test set, both containing $20$ images at resolution $565 \times 584$ pixels. The images are recorded with a fundus camera with $30^\circ$ field of view and are provided with manually labeled ground truth images from two different observers. Similar to existing works, we used the ground truth of the first observer as gold standard for the evaluation of our method.

%In Figure~\ref{fig:datasets}, we report example images from the three data sets, wogether with the corresponding ground truth images. 

\section{Experiments}
\label{sec:experiments}
\subsection{Evaluation}
For the rose stem and road crack center-line detection, we compute the precision (Pr), recall (Re) and F-score (F). 
%:
%\begin{equation}
%Pr = \frac{TP}{TP+FP}, Re = \frac{TP}{TP+FN}, F = \frac{2\cdot Pr \cdot Re}{Pr + Re}
%\label{eq:metrics}
%\end{equation}
For the computation of these metrics, we consider a certain amount of pixel distance $d^*$ to account for tolerance in the position of the detected center-line with respect to the position in the ground truth images~\cite{Zou2012}. We used $d^*=3$ for the TB-roses-1 data set and $d^*=2$ for the CrackTree206 data set (as reported in~\cite{Zou2012}). We compute the evaluation metrics by thresholding the output of RUSTICO with different values of threshold $t$, ranging from $0.01$ to $1$ in steps of $0.01$ and report the results for the value $t^*$ of the threshold that contributes to the highest average F-score on the considered data set. 

In the case of retinal vessel delineation, we evaluate the performance of the proposed method by computing the Matthews correlation coefficient (MCC) and the Connectivity-Area-Length (CAL) measure~\cite{Arias12}. The MCC is a reliable measure of accuracy for two-class classification problems where the cardinality of the classes is unbalanced. A value of $1$ indicates perfect classification, while $0$ and $-1$ correspond to random and completely wrong classification, respectively.

Pixel-wise comparison of the output of delineation algorithms with respect to ground truth images is subject to various problems. For instance, a displacement of the segmented image of one pixel in any direction would cause a substantial reduction of performance results. Moreover, evaluation of the performance against different ground truth images causes disparate results. The CAL was demonstrated to be robust to such issues and to be in accordance with perceptual quality of the segmentation output~\cite{Arias12}. 
The CAL measure is computed as the product of three measures of connectivity, area and length of the segmented line networks compared with the corresponding ground truths. Each of the single measures has values between 0 and $1$, with $0$ indicating complete difference, while $1$ representing a perfect match with the ground truth. 
For further details about the computation of the CAL metric we refer the reader to~\cite{Arias12}.

\subsection{Results and discussions}
\begin{figure}[!t]
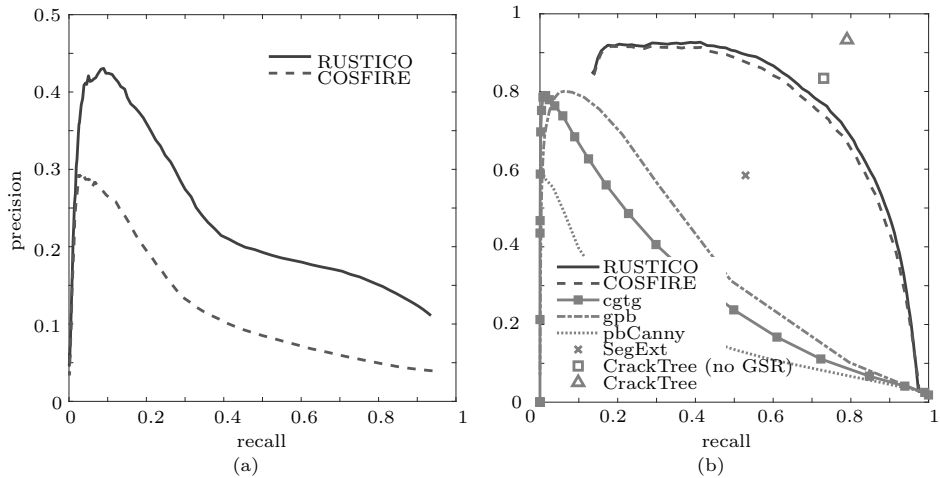

\scriptsize
\centering
\begin{tabular}{cc}
\input{rosecurve-paper.eps_tex} &
\input{crackcurve-paper.eps_tex} \\
(a) & (b)
\end{tabular}
 \caption{Precision-Recall curves achieved by RUSTICO (solid line) and COSFIRE (dashed line) on the (a) TB-roses-1 and (b) CrackTree data sets, together with the results achieved by existing approaches. }
\label{fig:curves}
\end{figure}

In Table~\ref{tab:results}, we report the results achieved by the proposed method, in comparison with those of the original COSFIRE filter. For the three considered applications, the push-pull inhibition mechanism that we embedded in the RUSTICO filter contributed to a substantial improvement of the delineation output. We achieved an increase of the value of the performance measures that is statistical significant (TB-roses: $p < 0.01$; CrackTree206: $p<0.01$: DRIVE: $p< 0.05$). The improvement of results is  evident from Fig.~\ref{fig:curves}, where we show 
the precision-recall curves achieved by the RUSTICO (solid line) and COSFIRE (dashed line) filters on the TB-roses-1 (Fig.~\ref{fig:curves}a) and the CrackTree206 (Fig.~\ref{fig:curves}b) data sets. The curves show a substantial  improvement of performance of RUSTICO with respect to COSFIRE in applications with images containing noise and spurious textures. For the CrackTree206 data set, we report the results achieved by existing methods: COSFIRE\cite{Caip17} ($F=0.6630$), SegExt ($F=0.55$), Canny ($F=0.26$), global pb ($F=0.44$) and pbCGTG ($F=0.35$). These results, except for those of the CrackTree~\cite{Zou2012} algorithm with and without pre-processing ($F=0.85$ and $F=0.77$, respectively), although specifically designed to deal with the characteristics of the concerned images, are considerably lower than those of RUSTICO ($F=0.6846$).  

\begin{table}[!t]
\caption{Comparison of the performance of the COSFIRE filter (C) and the proposed RUSTICO (R) on the TB-roses, CrackTree206 and DRIVE data sets. As quantitative measurements, for the former two data sets we use the F-score (F), while for the latter data set we use the CAL metric. The filters have different parameters according to the data set (TB-roses: $\sigma=2.5, \rho=16, \sigma_0=3, \alpha=0.1, \lambda=0.5, \xi=1.5$; CrackTree206: $\sigma=5.7, \rho=12, \sigma_0=5, \alpha=0.1, \lambda=3, \xi=2$: DRIVE: $\sigma=2.1, \rho=10, \sigma_0=3, \alpha=0.2, \lambda=3, \xi=1$). The parameters $\lambda$ and $\xi$ are specific of RUSTICO and not used for COSFIRE.}

\renewcommand{\arraystretch}{1}
  \centering
\footnotesize
\begin{tabular}{C{2cm}|C{1.5cm}C{1.5cm}|C{1.5cm}C{1.5cm}|C{1.5cm}C{1.5cm}}
\hline \hline
 & \multicolumn{2}{c}{\bfseries TB-roses-1} & \multicolumn{2}{c}{\bfseries CrackTree206} & \multicolumn{2}{c}{\bfseries DRIVE} \\ \hline
\textbf{\# images} & \multicolumn{2}{c}{35} & \multicolumn{2}{c}{206} & \multicolumn{2}{c}{20} \\ \hline \hline
~ & \multicolumn{2}{c|}{\bfseries F} & \multicolumn{2}{c|}{\bfseries F} & \multicolumn{2}{c}{\bfseries CAL} \\ \hline
~ & \bfseries C & \bfseries R & \bfseries C & \bfseries R & \bfseries C & \bfseries R \\ \hline
%$\bm{\sigma}$ & $2.5$ & $2.5$ & $5.7$ & $5.7$ & $2.1$ & $2.1$ \\
%$\bm{\rho}$ & $33$ & $33$ & $24$ & $24$ & $21$ & $21$ \\
%$\bm{\sigma_0}$ & $3$ & $3$ & $5$ & $5$ & $3$ & $3$ \\
%$\bm{\alpha}$ & $0.1$ & $0.1$ & $1.5$ & $1.5$ & $0.2$ & $0.2$  \\
%$\bm{\lambda}$ & - & $0.5$ & - & $3$ & - & $3$ \\
%$\bm{\xi}$ & - & $1.5$ & - & $2$ & - & $1$ \\ \hline \hline
 
\bfseries avg. & $0.3385$ & $\bm{0.3822}$ & $0.6630$ & $\bm{0.6846}$ & $0.7213$ & $\bm{0.7280}$ \\ \hline
\bfseries $\bm{p}$ & \multicolumn{2}{c|}{$\bm{<0.01}$} & \multicolumn{2}{c|}{$\bm{<0.01}$} & \multicolumn{2}{c}{$\bm{<0.05}$} \\
\hline \hline
\end{tabular}
\label{tab:results}
\end{table}

In Table~\ref{tab:compdrive}, we report the quantitative comparison of the results achieved by  RUSTICO with respect to those obtained by existing approaches on the DRIVE data set of retinal images. Among the methods based on filtering and mathematical morphology (above the middle line), RUSTICO achieves the best MCC and CAL measures with high statistical significance. We computed the performance of existing methods for which the segmentation output is publicly available. In the case of RORPO, we run experiments by varying the parameters with a grid search and reported the best obtained values. Methods based on machine learning perform generally better on pixel-wise classification, but show lower or comparable  performance  with that of RUSTICO in terms of CAL.

\begin{table}[!t]
\caption{Comparison of the results achieved by RUSTICO on the DRIVE data set with those obtained by existing method. The sign - indicates no statistical difference, while $*$ and $**$ indicate that the corresponding results are statistically higher than those of RUSTICO with significance level $0.05$ and $0.01$, respectively.}
	\renewcommand{\arraystretch}{1}
	\centering
	\begin{tabular}{l|C{2cm}C{2cm}|C{2cm}C{2cm}} \hline \hline
	\bfseries Method & \bfseries MCC & $\bm{p}$ & \bfseries CAL & $\bm{p}$ \\ \hline \hline
	RUSTICO & $\bm{0.7287}$ & - & $\bm{0.7280}$  & - \\ %
	COSFIRE~\cite{Azzopardi2015} & $0.7189$ & $\mathit{<0.01}$  & $0.7213$ & $\mathit{<0.05}$ \\   
	Jiang~\emph{et al.}~\cite{Jiang2003} & $0.6378$ & $\mathit{<0.01}$ & $0.5178$ & $\mathit{<0.01}$ \\	
	Perez~\emph{et al.}~\cite{MartinezPerez} & $0.6645$ & $\mathit{<0.05}$ & $0.5673$ & $\mathit{<0.01}$ \\
	
	RORPO~\cite{rorpoPAMI} & $0.6871$ & $\mathit{<0.01}$ & $0.6228$ & $\mathit{<0.01}$ \\ \hline
	Zana~\emph{et al.}~\cite{Zana2001} & $0.7258$ & - & $0.6180$ & $\mathit{<0.01}$ \\
	
	Staal~\emph{et al.}~\cite{StaalDrive2004} & $0.7378$ & $^{*}$ & $0.7010$ & $\mathit{<0.05}$ \\
	Niemeijer~\emph{et al.}~\cite{Niemeijer2004} & $0.7222$ & - & $0.6589$ & $\mathit{<0.01}$ \\
	FC-CRF~\cite{Orlando2016} & $\bm{0.7556}$ & $^{**}$ & $\bm{0.7311}$ & - \\ %%%%
	UP-CRF~\cite{Orlando2016} & $0.7401$ & $^{*}$ & $0.6747$ & $\mathit{<0.01}$ \\ \hline \hline
	
	\end{tabular}
	
	\label{tab:compdrive}
\end{table}

\section{Conclusions}
\label{sec:conclusions}

We presented a new method for the delineation of elongated patterns in images with spurious texture, named RUSTICO, that incorporates a push-pull inhibition mechanism operated by some neurons in area V1 of the visual cortex. RUSTICO takes input from two COSFIRE filters with opposite polarity and responds to elongated patterns also when they are surrounded by noise. 
We demonstrated how  the findings of  neuro-physiological studies of the visual system into image processing algorithms can be used to design more robust algorithms. The push-pull inhibition included in RUSTICO contributed to a statistically significant improvement with respect to existing methods in applications of delineation of rose stems for automatic gardening, detection of cracks in road surfaces and segmentation of blood vessels in medical images. We created and made available a data set of $100$ labeled images to test algorithm for segmentation of rose stems.

\section*{Acknowledgment}
This research received funding from the EU H2020 research and innovation framework (grant no. 688007, TrimBot2020).

\clearpage

\bibliographystyle{splncs}
\bibliography{egbib}
\end{document}